\documentclass[runningheads]{llncs}
\usepackage{xcolor}
\usepackage{amssymb}
\usepackage{booktabs}
\usepackage{multirow}
\usepackage{amsmath}
\usepackage{fontspec}
\usepackage{polyglossia}
\setotherlanguage{arabic}
\newcommand{\ar}[1]{{\arabicfont #1}}
\usepackage{graphicx}
\usepackage{booktabs}
\usepackage{array}

\usepackage{fontspec}
\usepackage{polyglossia}
\setmainlanguage{english}
\setotherlanguage{arabic}
\newfontfamily\arabicfont[ Script=Arabic]{Amiri-Regular.ttf}
\begin{document}

\title{Understanding Cross-Language Transfer Improvements in Low-Resource HTR: The Role of Sequence Modeling}
%
%
\author{Sana Al-azzawi\orcidID{0000-0001-7924-4953} \and Chang Liu\orcidID{0000-0002-7353-0251} \and Nudrat Habib\orcidID{0009-0005-9310-1701}
Elisa Barney\orcidID{0000-0003-2039-3844} \and
Marcus Liwicki\orcidID{0000-0003-4029-6574} }
\authorrunning{S. Al-azzawi et al.}

%
\date{}
\institute{
Luleå University of Technology\\
Department of Computer Science, Electrical and Space Engineering\\
Luleå, 97187, Sweden\\
\email{sana.al-azzawi@ltu.se, chang.liu@ltu.se, nudrat.habib@ltu.se, elisa.barney@ltu.se, marcus.liwicki@ltu.se}
}

\maketitle              
\begin{center}

Luleå University of Technology, Department of Computer Science, Electrical and Space Engineering , Luleå, Sweden

sana.al-azzawi@ltu.se, chang.liu@ltu.se, nudrat.habib@ltu.se, elisa.barney@ltu.se, marcus.liwicki@ltu.se

\end{center}
\begin{abstract}
Handwritten Text Recognition (HTR) for Arabic-script languages benefits from cross-language joint training under low-resource conditions, particularly when using CRNN-based models that combine convolutional encoders with sequence modeling. However, it remains unclear whether these improvements are better explained by shared visual representations or sequence-level dependencies. In this work, we conduct a controlled architectural study of line-level Arabic-script HTR, comparing CNN-only models with CTC decoding and CRNN models under identical single-script and multi-script training regimes. Experiments are performed on Arabic (KHATT), Urdu (NUST-UHWR), and Persian (PHTD) datasets under low-resource settings ($K \in {100, 500, 1000}$). Our results show a clear divergence in transfer behavior: while CNN-only models exhibit limited or unstable improvements, CRNN models achieve better performance under multi-script training, particularly in the most data-constrained regimes. Focusing on transfer improvements ($\Delta{CER}$) rather than absolute performance, we find that cross-language improvements are associated with sequence-level modeling, while sharing visual representations learned by the CNN encoder, corresponding to similarities in character shapes across scripts, alone appears to be insufficient. This finding suggests that contextual modeling plays an important role in enabling effective transfer in low-resource scenarios, and that similar behavior may extend to other low-resource language settings.

\end{abstract}
\section{Introduction}
Handwritten Text Recognition (HTR) is the task of converting images of handwritten text into machine-readable sequences. Recent advances in computer vision, natural language processing, and deep learning have significantly improved recognition accuracy, particularly for documents written in Latin-based scripts. In contrast, documents using other writing systems, including Arabic script, have received comparatively less attention and remain challenging, especially under limited data conditions~\cite{bilgin2023printed}.

cross-language training has recently emerged as an effective strategy for improving HTR in low-resource Arabic-script languages~\cite{al2026Cross}.
By leveraging data from related languages such as Arabic, Urdu, and Persian, joint training can substantially improve recognition performance when labeled data is scarce. However, it remains unclear
what explains these improvements in performance: are they better associated with shared visual representations (character shapes) or sequence-level modeling (e.g., character sequences) of structural dependencies?


In this work, we move beyond establishing the benefits of cross-language transfer and instead ask a more targeted question: \textbf{\emph{which architectural components are responsible for these improvements?}} To address this, we conduct a controlled architectural study, comparing CNN-only models with CTC decoding and CRNN models under identical single-script and multi-script training regimes. Rather than focusing on absolute performance, we analyze transfer improvements ($\Delta CER$) to better understand how auxiliary data is utilized across architectures. This comparison allows us to investigate how sequence-level modeling relates to to cross-language transfer behavior.

Prior work in scene text recognition has explored CNN-only architectures with CTC decoding as an alternative to recurrent models, mainly for efficiency and reduced model size~\cite{borisyuk2018rosetta,baek2019wrong}. Similar recurrence-free approaches have also been applied to handwritten text recognition~\cite{liu2020offline,coquenet2020recurrence}. In contrast, this work compares a CNN-only model with a CRNN model to investigate the role of sequence modeling and to investigate whether the benefits of cross-language transfer are better explained by visual similarity or by sequence-level modeling.

Experiments on Arabic (KHATT)~\cite{mahmoud2014khatt}, Urdu (NUST-UHWR)~\cite{ul2022convolutional}, and Persian (PHTD)~\cite{alaei2012dataset} under low-resource settings ($K \in \{100, 500, 1000\}$ text-line images) show a clear pattern: CNN-only models have limited or unstable improvements, while CRNN models consistently improve with cross-language training. This suggests that architectures incorporating sequence-level modeling play an important role in enabling effective cross-language transfer.
\newline
\newline
\textbf{Contributions.}
This paper makes the following contributions:
\begin{itemize}
    \item We present a controlled comparison between CNN-only and CRNN architectures, using the CNN-only model as a baseline to study the role of sequence modeling in cross-language transfer for Arabic-script HTR.
    
    \item We show that architectures incorporating sequence-level modeling are consistently associated with larger and more stable cross-language transfer improvements ($\Delta CER$), particularly under extreme low-resource conditions.
    
    \item We observe that CNN-only models, which rely on visual feature sharing, tend to yield more limited or less stable improvements in our setting compared to models with sequence modeling.

    
    \item We release code and experimental configurations to facilitate reproducibility.\footnote{GitHub link will be publicly released upon acceptance}
\end{itemize}

The remainder of this paper is organized as follows.
We first review related work in Section~\ref{sec:related}, followed by the proposed methodology in Section~\ref{sec:methodology}, which also describes the datasets and experimental setup used in our study. 
The quantitative, qualitative, and ablation results are presented and analyzed in Section~\ref{sec:results}.
Finally, Section~\ref{sec:conclusion} concludes the paper and outlines future directions.

\section{Background and Related Work}\label{sec:related}

\subsection{Arabic-script HTR and sequence modeling}

HTR for documents written in languages that use Arabic script, such as Arabic, Persian, and Urdu, remains challenging, particularly in low-resource settings where only limited labeled data is available. Although modern deep neural models achieve strong performance with sufficient training data, their accuracy degrades significantly when supervision is scarce~\cite{luo2020learn}.

These challenges are amplified by the properties of the Arabic writing system. The Arabic-script exhibits cursive connectivity, position-dependent character forms (e.g., \ar{ع} and \ar{ه} take different forms depending on position within a word, including initial, medial, and final forms), and a high degree of visual similarity between characters that differ only by dots or diacritics (e.g., \ar{خ}, \ar{ح}, \ar{ج})~\cite{salaheldin2025advancements}. Multiple languages use the Arabic script and share a largely common character inventory, with substantial overlap in base letterforms across scripts, as summarized in Table~\ref{tab:char_overlap}~\cite{ahmad2017impact}. This provides an opportunity.

Beyond script-level similarity, these languages are also historically and linguistically interconnected. Urdu contains a large proportion of Arabic-derived vocabulary, while Persian has borrowed many lexical items from Arabic over centuries~\cite{salaheldin2025advancements,nemati2022persian}. This makes Arabic-script languages well suited for cross-language learning.

Modern HTR systems~\cite{corbille2025applying,barrere2024training} combine convolutional encoders with sequence modeling, such as recurrent or attention-based models, to capture context along the text line. This helps resolve ambiguities from visually similar characters in handwritten text, including cursive handwriting.

Such convolutional-recurrent architectures have demonstrated strong performance across both Latin-script and Arabic-script HTR benchmarks, achieving competitive or state-of-the-art results in several studies~\cite{corbille2025applying,retsinas2022best,al2026cer}.

Despite the success of sequence-based models, several works in scene text recognition have explored CNN-only architectures with CTC decoding as an alternative to recurrent models, primarily to reduce computational complexity and memory consumption~\cite{borisyuk2018rosetta,baek2019wrong}. Similar recurrence-free approaches have also been applied to HTR including CNN-CTC models for Chinese HTR~\cite{liu2020offline} and fully convolutional architectures trained with CTC~\cite{coquenet2020recurrence}.

While these approaches demonstrate that recognition is possible without explicit sequence modeling, they do not clarify the role of sequence modeling in cross-language transfer.
In this work, we instead use a CNN-only model as a controlled baseline to study the contribution of sequence modeling, allowing us to assess its role in cross-language transfer.

\begin{table}[t]
\centering
\caption{Character inventory overlap across Arabic, Persian, and Urdu, derived from the datasets used in this study. Each row lists a set of characters, and a checkmark indicates their presence in the corresponding language. Note that the listed characters reflect only those observed in the datasets.}
\small
\begin{tabular}{p{6cm} c c c}
\toprule
\textbf{Characters} & \textbf{ Arabic } & \textbf{ Persian } & \textbf{ Urdu } \\
\midrule

\ar{ا ب ت ث ج ح خ د ذ ر ز ش س ص } & \checkmark & \checkmark & \checkmark \\
\ar{ض ط ظ ع غ ف ق ل م ن ه و ى ي} & \checkmark & \checkmark & \checkmark \\
\\
\ar{گ پ چ ژ ک} & -- & \checkmark & \checkmark \\
\\
\ar{ك} & \checkmark & \checkmark & -- \\
\\
\ar{ء} & \checkmark & -- & \checkmark \\
\\
\ar{ٹ ڈ ڑ ں ۂ ھ ڤ} & -- & -- & \checkmark \\
\\
\ar{إ} & \checkmark & -- & -- \\
\bottomrule
\end{tabular}
\label{tab:char_overlap}
\end{table}

\subsection{cross-language and low-resource HTR}

Multilingual and cross-script learning has been explored in both OCR and HTR. In particular, sequence-based models have been shown to generalize across languages within the same script family, indicating that shared structural patterns can be leveraged across languages~\cite{ul2013can}. More recently, cross-language training for Arabic-script HTR has demonstrated that joint training across related languages improves recognition performance under low-resource conditions~\cite{al2026Cross}.

Several works have also addressed low-resource HTR through data augmentation, synthetic data generation, or transfer learning. For example, synthetic data has been used to improve recognition for Ottoman Turkish~\cite{bilgin2023printed}, while other studies combine multiple datasets within the same language to increase training data~\cite{riaz2022conv,hamza2024network}. However, these approaches primarily focus on improving performance rather than analyzing the mechanisms underlying transfer.

Overall, prior studies either operate in printed OCR, high-resource settings, or single auxiliary–target configurations, and typically report only aggregate performance.

The question of why cross-language transfer works—specifically, whether it is associated with shared visual representations or sequence-level modeling—remains largely unexplored.

\section{Methodology}\label{sec:methodology}

\subsection{Problem Formulation}

We study cross-language transfer in Arabic-script HTR under controlled architectural conditions. 
Let $\mathcal{D}_t$ denote a target-language dataset and $\mathcal{D}_a$ auxiliary datasets drawn from related Arabic-script languages. 
Given a limited labeled subset $\mathcal{D}_t^{(K)}$ with $K \in \{100, 500, 1000\}$, we assume a joint training setup in which the model is trained on $\mathcal{D}_t^{(K)} \cup \mathcal{D}_a$ and evaluated on the target dataset.

Prior work has shown that cross-language transfer improves recognition performance under low-resource conditions~\cite{al2026Cross}. In this work, we move beyond establishing these improvements and instead aim to better understand which components of the recognition model contribute to this effect.

To this end, we consider two model families: a CNN-only model ($\mathcal{M}_{\text{CNN}}$) that captures visual representations, and a CRNN model ($\mathcal{M}_{\text{CRNN}}$) that additionally incorporates sequence modeling.
By comparing $\mathcal{M}_{\text{CNN}}$ and $\mathcal{M}_{\text{CRNN}}$ under identical training conditions, we aim to investigate the contribution of sequence modeling to cross-language transfer.

All experiments are conducted under a unified training and evaluation protocol to ensure a fair comparison between model architectures.

\subsection{Architectural Comparison}
\begin{figure}[t]
\centering
\includegraphics[width=0.75\textwidth]{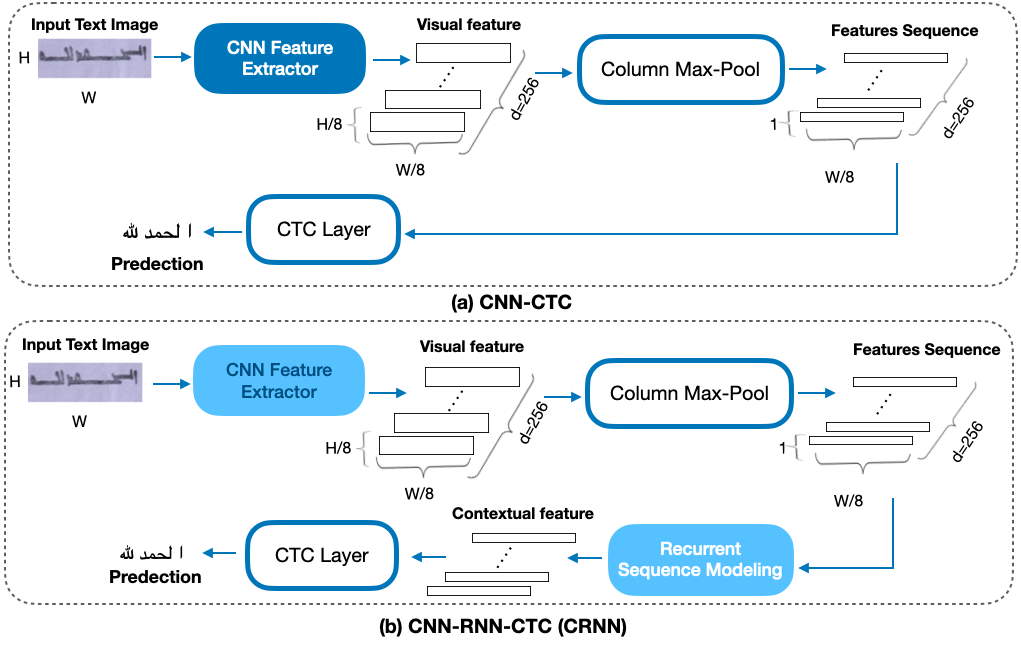}
\caption{Architectural comparison of the two model families used in this study. 
(a) CNN-CTC (CNN-only) model ($\mathcal{M}_{\text{CNN}}$), which relies solely on visual feature extraction followed by CTC decoding. 
(b) CNN-RNN-CTC (CRNN) model ($\mathcal{M}_{\text{CRNN}}$), which augments the visual encoder with recurrent sequence modeling before CTC prediction.}\label{fig:models}
\end{figure} 

To investigate the source of cross-language transfer improvements, we compare two model families under identical training conditions: a CNN-only model ($\mathcal{M}_{\text{CNN}}$) and a CRNN model ($\mathcal{M}_{\text{CRNN}}$).
As illustrated in Fig.~\ref{fig:models}, the two architectures differ only in the presence of recurrent sequence modeling.
\paragraph{\textbf{CNN-only model ($\mathcal{M}_{\text{CNN}}$).}}
The CNN-only model (Fig.~\ref{fig:models}a) consists of a convolutional feature extractor followed by a CTC prediction layer. 
The feature extractor maps the input image into a sequence of visual features, which are then used to predict character probabilities at each time step.

We employ a ResNet-based architecture as the feature extractor. 
It begins with an initial $7\times7$ convolutional layer with stride $2\times2$, followed by multiple residual blocks with increasing channel dimensions (64, 128, and 256). 
Each residual block contains two $3\times3$ convolutional layers with batch normalization and ReLU activation, together with identity shortcut connections. 
Max-pooling layers are used to progressively reduce spatial resolution.

To obtain a sequential representation, the resulting feature maps are collapsed along the vertical dimension using column-wise pooling, producing a one-dimensional feature sequence along the image width.

For prediction, we use CTC, which enables sequence prediction without explicit alignment between input features and output labels. 
At each time step, the model predicts character probabilities, and the final sequence is obtained by removing repeated characters and blank symbols.

This architecture relies solely on shared visual representations across scripts, without explicit sequence modeling.

\paragraph{\textbf{CRNN model ($\mathcal{M}_{\text{CRNN}}$).}}
The CRNN model (Fig.~\ref{fig:models}b) extends $\mathcal{M}_{\text{CNN}}$ by incorporating a sequence modeling component on top of the same convolutional feature extractor. 
While the CNN encoder produces a sequence of visual features, these features may lack sufficient contextual information for accurate character prediction.

To address this, the CRNN introduces a recurrent sequence modeling module consisting of three stacked bidirectional LSTM layers, each with 256 hidden units per direction. 
These layers model contextual dependencies along the sequence, allowing the model to capture long-range relationships between characters.

The output of the final BiLSTM layer is projected onto the character space through a fully connected layer, and prediction is performed using the same CTC objective as in $\mathcal{M}_{\text{CNN}}$.

We note that the CRNN model differs from the CNN-only model not only in the presence of sequence modeling, but also in increased representational capacity due to additional recurrent layers. As a result, this comparison reflects a combined architectural effect rather than attributing improvements solely to sequence modeling. Nevertheless, since both models share the same convolutional backbone and training protocol, the comparison provides useful insight into the role of contextual modeling in cross-language transfer.

\paragraph{Controlled comparison.}
Both $\mathcal{M}_{\text{CNN}}$ and $\mathcal{M}_{\text{CRNN}}$ share the same convolutional backbone, training procedure, and output vocabulary. 
The primary architectural difference is the inclusion of recurrent sequence modeling in $\mathcal{M}_{\text{CRNN}}$. 
This controlled setup allows us to examine how sequence-level modeling relates cross-language transfer behavior.

The CNN-only model ($\mathcal{M}_{\text{CNN}}$) contains approximately 5.8M parameters, while the CRNN model ($\mathcal{M}_{\text{CRNN}}$) contains 10.1M parameters due to the additional recurrent layers. 

To control for model capacity, we additionally evaluate a higher-capacity CNN-only variant ($\mathcal{M}_{\text{CNN-expand}}$) by increasing the width of the convolutional layers while keeping the overall architecture unchanged. Specifically, the channel dimensions are increased from (64, 128, 256) to (86, 172, 344), resulting in a model with approximately 10.4M parameters, closely matching the capacity of the CRNN.

This design enables a fairer comparison without altering the model topology, and allows us to assess whether the observed transfer improvement can be explained by model capacity alone.

\subsection{Training Setting}

For each target language, we consider both single-script and multi-script training settings. Let $J$ denote the number of auxiliary datasets in $\mathcal{D}_a$.

\paragraph{Single-script training ($J=0$).}
The model is trained using only $K$ labeled lines from the target dataset $\mathcal{D}_t^{(K)} $.

\paragraph{Multi-script training ($J=2$).}
The model is trained on the union of the limited target subset $\mathcal{D}_t^{(K)}$ and the auxiliary datasets $\mathcal{D}_a = \bigcup_{j=1}^{J} \mathcal{D}_a^{(j)}$:
\begin{equation}
\mathcal{D}_{train} = \mathcal{D}_t^{(K)} \cup \mathcal{D}_a.
\end{equation}
This setting reflects a realistic scenario in which annotated data for the target language is scarce, while related-script datasets are available. For each model ($\mathcal{M}_{\text{CNN}}$ and $\mathcal{M}_{\text{CRNN}}$), experiments are conducted across all values of $K$ under both single-script and multi-script training settings, following the protocol in~\cite{al2026Cross}. We refer the reader to that work for further details on the training configuration.

The effect of different auxiliary dataset configurations (e.g., using one auxiliary dataset ($J=1$) versus multiple auxiliary datasets ($J>1$)) has been investigated in prior work~\cite{al2026Cross}, where combining multiple auxiliary datasets was generally found to yield stronger improvements. In this work, we do not revisit this aspect and instead fix the multi-script setting ($J=2$) in order to focus on the effect of architectural differences on cross-language transfer improvements. Importantly, our goal is not to optimize auxiliary data composition, but to study how different architectures translate auxiliary data into transfer improvements under a consistent training setup.)

\subsection{Datasets}
We conduct experiments on three Arabic-script HTR datasets: KHATT (Arabic)~\cite{mahmoud2014khatt}, NUST-UHWR (Urdu)~\cite{ul2022convolutional}, and PHTD (Persian)~\cite{alaei2012dataset}. 
These datasets represent three distinct but closely related writing systems sharing a common script, enabling controlled evaluation of cross-language transfer.

KHATT is used as the Arabic benchmark, NUST-UHWR provides a large-scale Urdu dataset with diverse handwriting styles, and PHTD represents a low-resource Persian dataset with limited training data and reduced variability. PHTD serves as the most data-constrained setting, as it contains substantially fewer training samples (1,473 lines) ~\cite{al2026cer} and reduced textual diversity compared to KHATT and NUST-UHWR, allowing us to analyze transfer behavior under extreme low-resource conditions.

The original PHTD dataset~\cite{alaei2012dataset} is provided at the page level without predefined line segmentation or official splits. In this work, we use the line-level version and data partitioning introduced in~\cite{al2026cer}, which enables consistent training and evaluation.

All experiments are conducted at the line level, and evaluation is performed independently on each target dataset. Additional details on dataset statistics can be found in~\cite{al2026cer}.

Representative text line samples from these datasets are shown  with the results in Figure~\ref{fig:qualitative}.


\subsection{Implementation Details}

Models were implemented in PyTorch and trained on an NVIDIA A100 GPU. 
Optimization was performed using AdamW (batch size 16, initial learning rate $5\times10^{-4}$) with a multi-step learning rate schedule reducing the rate by 0.1 at 50\% and 75\% of training.

We first compute the average height and width of input line images across the combined training sets of KHATT, NUST-UHWR, and PHTD. Based on these statistics, we select a fixed input size of 110 pixels in height and 1450 pixels in width as representative values.

Input line images are converted to grayscale and rescaled to this fixed height while preserving aspect ratio, thereby upscaling smaller images and downscaling larger ones. The corresponding width is adjusted proportionally; if it exceeds 1450 pixels, the image is further rescaled to fit within this limit. 

Additionally, fixed horizontal padding of 64 pixels was applied to both sides of each image following~\cite{corbille2025applying}, which has been shown to improve recognition performance in line-level HTR. To align with left-to-right sequence modeling while handling right-to-left scripts, ground-truth transcriptions were reversed during training.

To improve robustness under low-resource conditions, we applied standard data augmentations, including affine transformations, elastic and grid distortions, morphological operations, and brightness and contrast adjustments. Augmentations were applied stochastically across datasets.

For each value of $K$, training was performed for a fixed number of optimization steps (iterations). 
In the multi-script setting, optimization was performed on $\mathcal{D}_{train} = \mathcal{D}_t^{(K)} \cup \mathcal{D}_a$, while validation and model selection were conducted exclusively on the target validation split.

To verify robustness, we repeated experiments with three different random seeds. While the figures report results from a single representative run for clarity, we observed consistent trends across seeds, with similar relative improvements between single-script and multi-script training.

Evaluation is performed using Character Error Rate (CER), the primary metric in this work, computed as the normalized edit distance between predicted and ground-truth transcriptions. 
The checkpoint with the lowest validation CER on the target dataset was selected for evaluation.

\section{Results}\label{sec:results}

This section presents the evaluation and analysis of the proposed architectural comparison, focusing on understanding the behavior of cross-language transfer improvements ($\Delta CER$) in Arabic-script HTR. We analyze how the CNN-only ($\mathcal{M}_{\text{CNN}}$) and the CRNN ($\mathcal{M}_{\text{CRNN}}$) models benefit from multi-script training under identical experimental conditions.

We emphasize that our analysis focuses on transfer changes ($\Delta$CER), defined as $\Delta$CER = $\mathrm{CER}{\text{Multi}} - \mathrm{CER}{\text{Single}}$, where negative values indicate improvement from multi-script over single-script training, rather than absolute CER values.

\subsection{CNN-only vs.\ CRNN Models under cross-language Training}

\begin{figure}[t]
\centering
\includegraphics[width=1\textwidth]{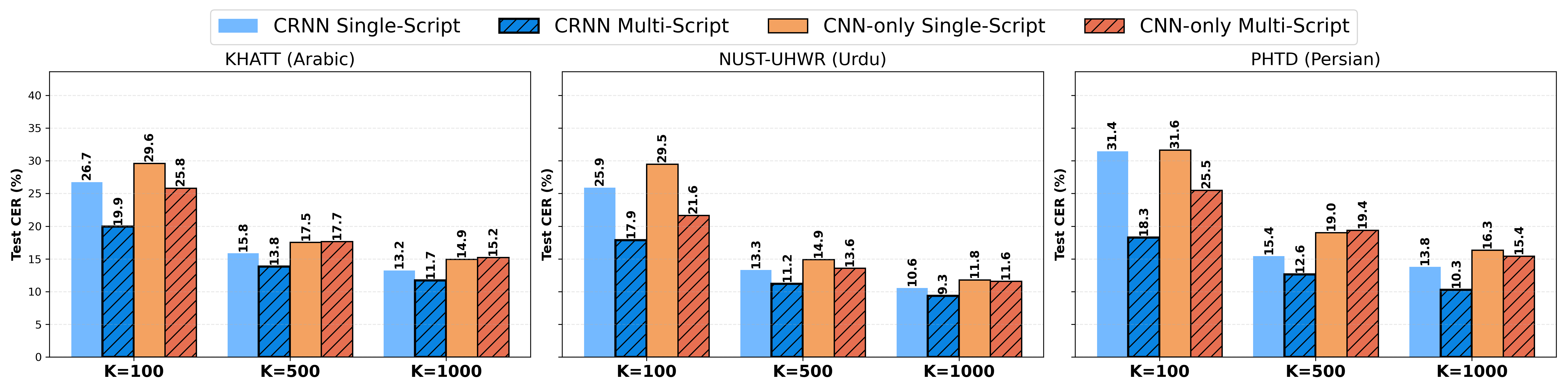}
\caption{Comparison of the CNN-only model ($\mathcal{M}_{\text{CNN}}$) and CRNN model ($\mathcal{M}_{\text{CRNN}}$) performance under single-script ($J{=}0$) and multi-script ($J{=}2$) training across low-resource regimes ($K{=}100, 500, 1000$), reported using CER (\%). Results are shown for KHATT (Arabic), NUST-UHWR (Urdu), and PHTD (Persian). Lower CER indicates better recognition performance.}\label{fig:crnn_vs_cnn}
\end{figure}

Figure~\ref{fig:crnn_vs_cnn} reports the behavior of the CNN-only and CRNN architectures under single-script ($J{=}0$) and multi-script ($J{=}2$) training.
Both architectures benefit from cross-language training, but they differ in how effectively they translate auxiliary data into transfer changes (ΔCER).
The CRNN consistently achieves larger and more stable improvements across datasets and data regimes. For example, on KHATT, the CRNN improves from 26.7 to 19.9 CER ($\Delta$CER = $-6.8$) at $K=100$, whereas the CNN-only achieves a smaller reduction in CER from 29.6 to 25.8 ($\Delta$CER = $-3.8$). A similar trend is observed on PHTD, where the CRNN achieves substantial improvements (e.g., 31.4 to 18.3 at $K=100$, $\Delta$CER = $-13.1$), while the CNN-only model improvements are more limited (31.6 to 25.5, $\Delta$CER = $-6.1$).

On NUST-UHWR, both architectures benefit strongly from multi-script training at $K=100$, with the CRNN improving from 25.9 to 17.9 CER and the CNN-only model from 29.5 to 21.6. However, as the number of target labeled samples increases, the improvements from multi-script training decrease. At $K=500$, the CRNN achieves 11.2 CER under multi-script training compared to 13.6 for the CNN-only model, and at $K=1000$, it achieves 9.3 compared to 11.6.

Overall, these results suggest that cross-language transfer is not fully explained by visual features alone, and is more consistently observed in models that incorporate sequence-level contextual modeling.




 \subsection{Analysis of Cross-Script Transfer Improvements}

\begin{figure}[t]
\centering
\includegraphics[width=1\textwidth]{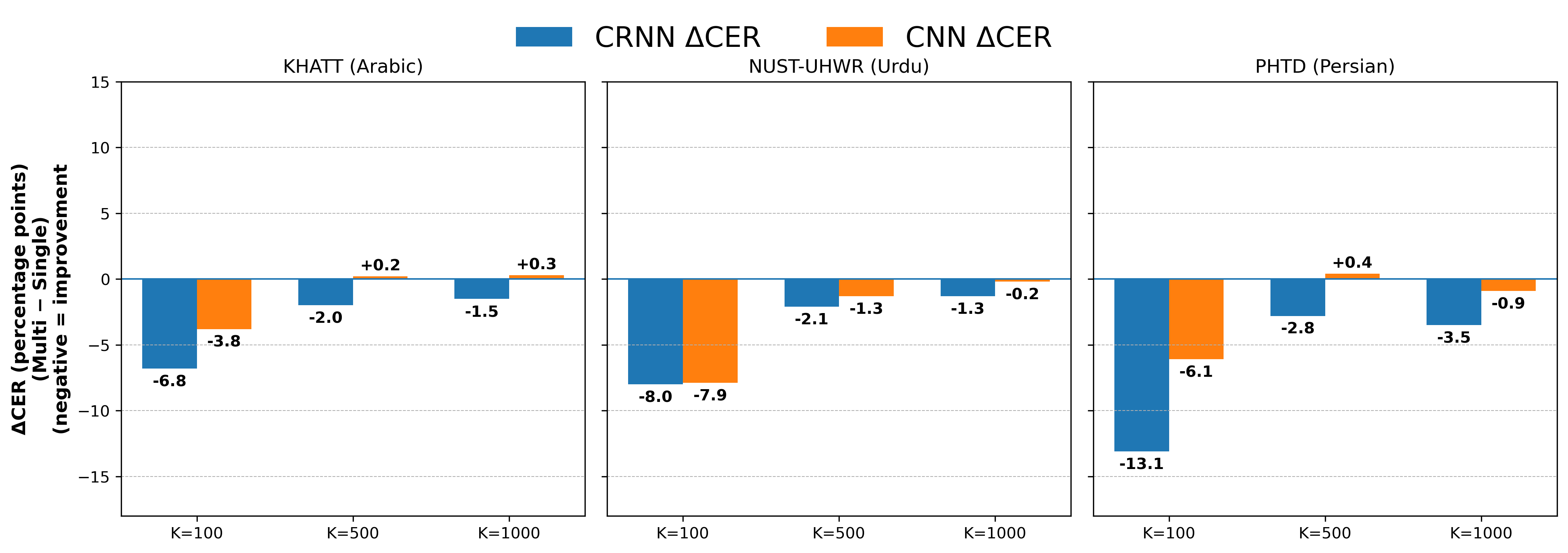}
\caption{cross-language transfer changes ($\Delta$CER = $\mathrm{CER}{\text{Multi}} - \mathrm{CER}{\text{Single}}$, in percentage points) for the CRNN and the CNN-only architectures across low-resource regimes ($K{=}100, 500, 1000$). Negative values indicate improvement from multi-script training over single-script training. The CRNN consistently achieves larger and more stable improvements, while the CNN-only exhibits weaker and sometimes positive $\Delta$CER values, indicating negative transfer.}\label{fig:improvment}
\end{figure}

Figure~\ref{fig:improvment} shows cross-language transfer changes ($\Delta$CER) for the CNN-only and the CRNN models across three low-resource regimes. 

First, the CRNN achieves larger improvements than the CNN-only, especially in low-resource settings. At $K{=}100$, CRNN improvements are substantially larger across all datasets, reaching -13.10 CER points on PHTD and -8.00 on NUST-UHWR, compared to more modest improvements from the CNN-only model (-6.10 and -7.90, respectively). This suggests that a substantial portion of the transfer benefit in extremely low-resource regimes is associated with shared sequence modeling rather than shared visual features alone.

Second, the CRNN transfer improvements decrease smoothly as $K$ increases, reflecting diminishing returns from auxiliary scripts as more target-script data becomes available. 

On KHATT, CRNN improvements decrease from -6.8 at $K{=}100$ to -2.0 and -1.5 at $K{=}500$ and $K{=}1000$, respectively, while the CNN-only shows unstable behavior, including negative transfer at higher data regimes ($+0.2$ and $+0.3$). In contrast, the CNN-only model exhibits less stable behavior, including negative transfer in some cases (e.g., PHTD at $K{=}500$).

Finally, the contrast between architectures is particularly pronounced on PHTD, where the CRNN achieves consistently strong improvements across all regimes (-13.10, -2.80, and -3.50), while the CNN-only improvements remain smaller and less consistent (-6.10, $+0.40$, and -0.90). On NUST-UHWR, both architectures benefit at $K{=}100$, with the CNN-only model achieving a comparable improvement (-7.90 vs.\ -8.00), but the CRNN maintains stronger improvements as $K$ increases (-2.10 vs.\ -1.30 at $K{=}500$, and -1.30 vs.\ -0.20 at $K{=}1000$).

Overall, these results provide consistent evidence that cross-language transfer in HTR is associated with sequence-level modeling, particularly in terms of the magnitude and stability of the observed improvements. This suggests that contextual modeling enables more effective use of auxiliary data when learning cross-language representations.

These observations may be influenced by differences in model capacity. To further examine whether these differences can be explained by model capacity, we evaluate a higher-capacity CNN-only baseline ($\mathcal{M}_{\text{CNN-expand}}$) with a parameter count similar to $\mathcal{M}_{\text{CRNN}}$. This is achieved by increasing the width of the convolutional layers while keeping the overall architecture unchanged, allowing for a fair comparison without altering the model topology.

We conduct this analysis on two datasets (KHATT and PHTD), which exhibit the largest differences in transfer improvement between the CNN-only ($\mathcal{M}_{\text{CNN}}$) and CRNN ($\mathcal{M}_{\text{CRNN}}$) models, thereby providing the most informative settings for this comparison.

\begin{table}[t]
\centering
\caption{Ablation study comparing cross-language transfer improvements ($\Delta$CER ) for the original CNN-only model ($\mathcal{M}_{\text{CNN}}$), a higher-capacity CNN-only model ($\mathcal{M}_{\text{CNN-expand}}$), and the CRNN model ($\mathcal{M}_{\text{CRNN}}$). Negative values indicate improvement from multi-script training over single-script training.}
\label{tab:cnn_ablation}
\begin{tabular}{lcccc}
\toprule
\textbf{Dataset} & \textbf{K} & \textbf{$\mathcal{M}_{\text{CNN}}$} & \textbf{$\mathcal{M}_{\text{CNN-expand}}$} & \textbf{$\mathcal{M}_{\text{CRNN}}$} \\
\midrule
\multirow{3}{*}{KHATT} 
& 100  & -3.80 & -2.58 & -6.80 \\
& 500  & +0.20 & +0.23 & -2.00 \\
& 1000 & +0.30 & +0.80 & -1.50 \\
\midrule
\multirow{3}{*}{PHTD} 
& 100  & -6.10 & -7.21 & -13.10 \\
& 500  & +0.40 & +1.34 & -2.80 \\
& 1000 & -0.90 & +0.83 & -3.50 \\
\bottomrule
\end{tabular}
\end{table}

Table~\ref{tab:cnn_ablation} shows the transfer improvements ($\Delta$CER) for the original CNN-only model and the higher-capacity CNN model. Overall, increasing the capacity of the CNN-only model does not lead to more consistent transfer benefits nor does it reach the performance improvements achieved by the CRNN. While a small benefit is observed for PHTD at $K=100$, the higher-capacity CNN ($\mathcal{M}_{\text{CNN-expand}}$) shows weaker or even positive $\Delta$CER values (indicating degradation) at higher data regimes. For example, on PHTD at $K=1000$, the $\Delta$CER changes from $-0.90$ to $+0.83$. A similar trend is observed on KHATT, where increasing model capacity often reduces benefits rather than improving them.

These findings suggest that increasing model capacity alone is not sufficient to explain the observed transfer performance improvements, and that architectural components such as sequence-level modeling play an important role in effectively leveraging auxiliary data, although this relationship should be interpreted in terms of association rather than strict causation.

\paragraph{Architectural takeaway.}
Figures~\ref{fig:crnn_vs_cnn} and~\ref{fig:improvment} show that both models benefit from cross-language training, but the CRNN achieves larger and more consistent improvements than the CNN-only model. The CNN-only model shows its strongest improvements in the most data-scarce regime, while the CRNN continues to improve as $K$ increases, suggesting that architectures incorporating sequence-level modeling play an important role in associated with more consistent transfer improvements.

\subsection{Qualitative Analysis}
\begin{figure}[h]
\centering
\includegraphics[width=\textwidth]{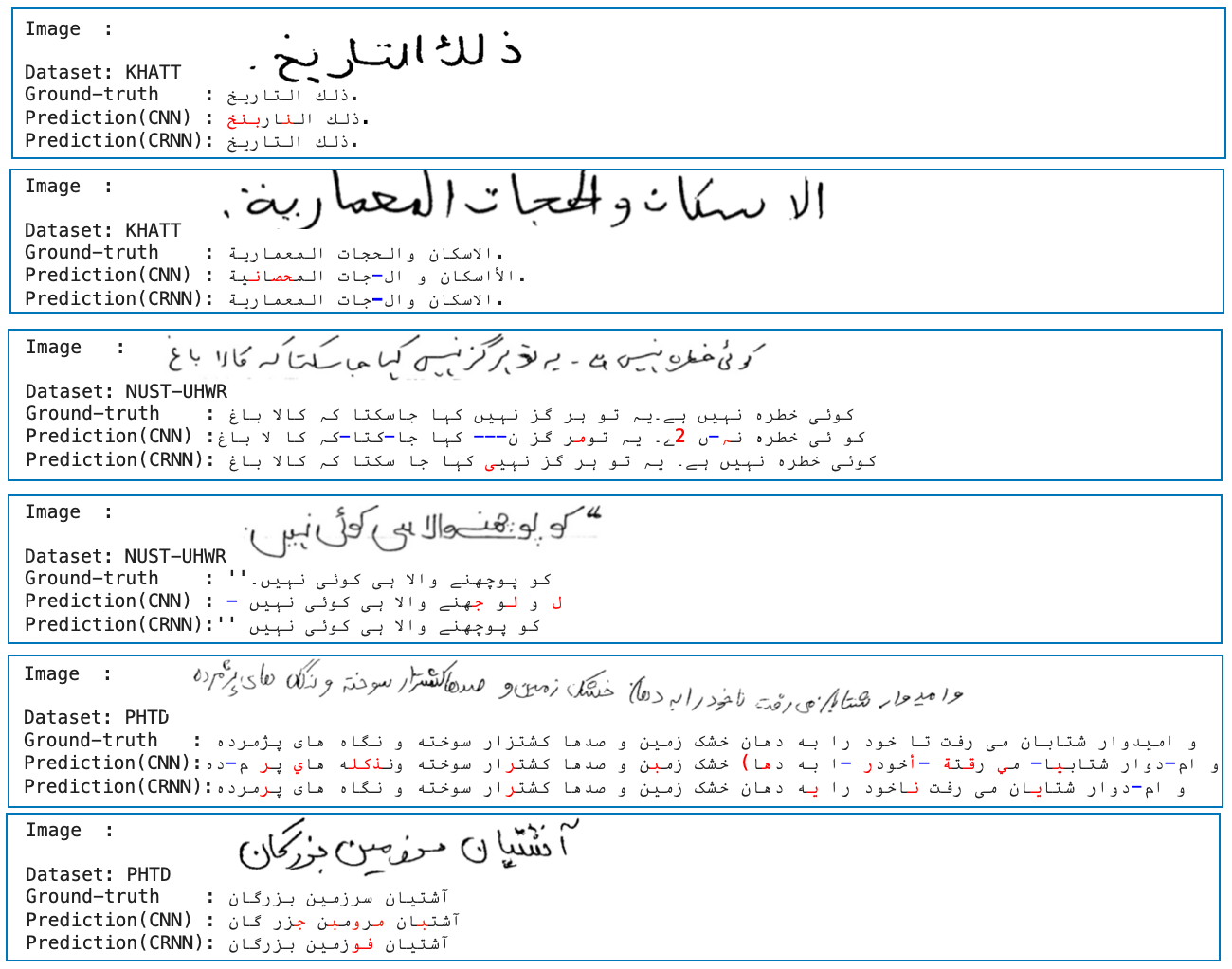}
\caption{Qualitative comparison of predictions from the CNN-only ($\mathcal{M}_{\text{CNN}}$) and CRNN ($\mathcal{M}_{\text{CRNN}}$) models on KHATT, NUST, and PHTD datasets under multi-script training ($J{=}2$). For each dataset, two representative examples are shown. Incorrect predicted characters are highlighted in red, while missing characters are indicated in blue. The examples illustrate differences in how the two architectures leverage contextual information when benefiting from cross-language training.}\label{fig:qualitative}
\end{figure} 

For qualitative analysis, Figure~\ref{fig:qualitative} shows six text-line samples and their corresponding ground-truth and predicted transcriptions for both the CNN-only and CRNN models under multi-script training ($J=2$). The examples are shown at $K=500$, as a representative setting. At $K=100$, predictions are often noisy, while at $K=1000$, differences between architectures are less pronounced.

Under multi-script training, the CNN-only model still exhibits errors such as missing characters and confusion between visually similar letters, whereas the CRNN model more consistently resolves these ambiguities. These examples illustrate how sequence-level modeling is associated with more coherent predictions when leveraging cross-language information, supporting the observed differences in transfer improvements ($\Delta$CER). While qualitative examples are shown under multi-script training, they reflect the same trends observed in the quantitative analysis across architectures.

\section{Conclusion}\label{sec:conclusion}

In this work, we investigated the source of cross-script transfer improvements in low-resource Arabic-script HTR through a controlled architectural comparison between CNN-only and CRNN models. Rather than focusing on absolute performance, our analysis examined how different architectures benefit from cross-language training. We found that architectures incorporating sequence-level modeling are associated with more consistent and stable cross-language transfer improvements across related scripts. While CNN-only models can capture shared visual features, they tend to yield more limited and sometimes unstable improvements under multi-script training. In contrast, models with sequence-level modeling more reliably leverage auxiliary data to achieve transfer improvements, particularly in the most data-constrained settings. Importantly, while both architectures benefit from cross-language training, the CNN-only model shows its strongest improvements in the most data-scarce regime, whereas the CRNN achieves larger and more consistent improvements across data regimes. These findings suggest that cross-language transfer improvements are consistently larger in architectures that incorporate sequence-level modeling, although visual feature sharing alone does not consistently lead to stable transfer improvements in our experiments. Future work will explore extending this analysis to transformer-based architectures and studying cross-language transfer across more diverse script families, including transfer between Latin and Arabic scripts.


 \section{Acknowledgements} 
This research is financially supported by the European Regional Development Fund and the MARTINA-project (no. 20367152) and partially supported by the Wallenberg AI, Autonomous Systems and Software Program (WASP).

\bibliography{ref}

@article{bilgin2023printed,
  title={{Printed Ottoman text recognition using synthetic data and data augmentation}},
  author={Bilgin Tasdemir, Esma F},
  journal={International Journal on Document Analysis and Recognition (IJDAR)},
  volume={26},
  number={3},
  pages={273--287},
  year={2023},
  publisher={Springer}
}

@article{al2026Cross,
  title={{Cross-Language Learning within Arabic Script for Low-Resource HTR}},
  author={Al-azzawi, Sana and Barney, Elisa and Liwicki, Marcus},
  journal={arXiv:2605.02089},
  year={2026}
}

@article{al2026cer,
  title={{CER-HV: A human-in-the-loop framework for cleaning datasets applied to Arabic-script HTR}},
  author={Al-azzawi, Sana and Barney, Elisa and Liwicki, Marcus},
  journal={arXiv preprint arXiv:2601.16713},
  year={2026}
}

@article{mahmoud2014khatt,
  title={{KHATT: An open Arabic offline handwritten text database}},
  author={Mahmoud, Sabri A and Ahmad, Irfan and Al-Khatib, Wasfi G and Alshayeb, Mohammad and Parvez, Mohammad Tanvir and M{\"a}rgner, Volker and Fink, Gernot A},
  journal={Pattern Recognition},
  volume={47},
  number={3},
  pages={1096--1112},
  year={2014},
  publisher={Elsevier}
}

@article{ul2022convolutional,
  title={{A convolutional recursive deep architecture for unconstrained Urdu handwriting recognition}},
  author={ul Sehr Zia, Noor and Naeem, Muhammad Ferjad and Raza, Syed Muhammad Kumail and Khan, Muhammad Mubasher and Ul-Hasan, Adnan and Shafait, Faisal},
  journal={Neural Computing and Applications},
  volume={34},
  number={2},
  pages={1635--1648},
  year={2022},
  publisher={Springer}
}

@article{alaei2012dataset,
  title={Dataset and ground truth for handwritten text in four different scripts},
  author={Alaei, Alireza and Pal, Umapada and Nagabhushan, P},
  journal={International Journal of Pattern Recognition and Artificial Intelligence},
  volume={26},
  number={04},
  pages={1253001},
  year={2012},
  publisher={World Scientific}
}

@inproceedings{corbille2025applying,
  title={Applying Center Loss to Neural Networks for Sequence Prediction: A Study for Handwriting Recognition},
  author={Corbill{\'e}, Simon and Barney Smith, Elisa H},
  booktitle={International Joint Conference on Neural Networks (IJCNN)},
  year={2025},
  organization={IEEE}
}

@inproceedings{baek2019wrong,
  title={What is wrong with scene text recognition model comparisons? dataset and model analysis},
  author={Baek, Jeonghun and Kim, Geewook and Lee, Junyeop and Park, Sungrae and Han, Dongyoon and Yun, Sangdoo and Oh, Seong Joon and Lee, Hwalsuk},
  booktitle={{Proceedings of the IEEE/CVF International Conference on Computer Vision}},
  pages={4715--4723},
  year={2019}
}

@inproceedings{borisyuk2018rosetta,
  title={Rosetta: Large scale system for text detection and recognition in images},
  author={Borisyuk, Fedor and Gordo, Albert and Sivakumar, Viswanath},
  booktitle={{Proceedings of the 24th ACM SIGKDD International Conference on Knowledge Discovery \& Data Mining}},
  pages={71--79},
  year={2018}
}

@article{liu2020offline,
  title={{Offline handwritten Chinese text recognition with convolutional neural networks}},
  author={Liu, Brian and Xu, Xianchao and Zhang, Yu},
  journal={arXiv preprint arXiv:2006.15619},
  year={2020}
}

@inproceedings{coquenet2020recurrence,
  title={Recurrence-free unconstrained handwritten text recognition using gated fully convolutional network},
  author={Coquenet, Denis and Chatelain, Cl{\'e}ment and Paquet, Thierry},
  booktitle={2020 17th International Conference on Frontiers in Handwriting Recognition (ICFHR)},
  pages={19--24},
  year={2020},
  organization={IEEE}
}

@inproceedings{retsinas2022best,
  title={Best practices for a handwritten text recognition system},
  author={Retsinas, George and Sfikas, Giorgos and Gatos, Basilis and Nikou, Christophoros},
  booktitle={International Workshop on Document Analysis Systems},
  pages={247--259},
  year={2022},
  organization={Springer}
}

@inproceedings{luo2020learn,
  title={Learn to augment: Joint data augmentation and network optimization for text recognition},
  author={Luo, Canjie and Zhu, Yuanzhi and Jin, Lianwen and Wang, Yongpan},
  booktitle={{Proceedings of the IEEE/CVF Conference on Computer Vision and Pattern Recognition}},
  pages={13746--13755},
  year={2020}
}

@article{salaheldin2025advancements,
  title={{Advancements and challenges in Arabic optical character recognition: A comprehensive survey}},
  author={Salaheldin Kasem, Mahmoud and Mahmoud, Mohamed and Kang, Hyun-Soo},
  journal={ACM Computing Surveys},
  volume={58},
  number={4},
  pages={1--37},
  year={2025},
  publisher={ACM New York, NY}
}

@inproceedings{ahmad2017impact,
  title={{The impact of visual similarities of Arabic-like scripts regarding learning in an OCR system}},
  author={Ahmad, Riaz and Naz, Saeeda and Afzal, M Zeshan and Rashid, S Faisal and Liwicki, Marcus and Dengel, Andreas},
  booktitle={2017 14th IAPR International Conference on Document Analysis and Recognition (ICDAR)},
  volume={7},
  pages={15--19},
  year={2017},
  organization={IEEE}
}

@inproceedings{ul2013can,
  title={{Can we build language-independent OCR using LSTM networks?}},
  author={Ul-Hasan, Adnan and Breuel, Thomas M},
  booktitle={Proceedings of the 4th International Workshop on Multilingual OCR},
  pages={1--5},
  year={2013}
}

@article{riaz2022conv,
  title={{Conv-transformer architecture for unconstrained off-line Urdu handwriting recognition}},
  author={Riaz, Nauman and Arbab, Haziq and Maqsood, Arooba and Nasir, Khuzaeymah and Ul-Hasan, Adnan and Shafait, Faisal},
  journal={International Journal on Document Analysis and Recognition (IJDAR)},
  volume={25},
  number={4},
  pages={373--384},
  year={2022},
  publisher={Springer}
}

@article{hamza2024network,
  title={{ET-Network: A novel efficient transformer deep learning model for automated Urdu handwritten text recognition}},
  author={Hamza, Ameer and Ren, Shengbing and Saeed, Usman},
  journal={{PLOS One}},
  volume={19},
  number={5},
  pages={e0302590},
  year={2024},
  publisher={Public Library of Science San Francisco, CA USA}
}

@article{nemati2022persian,
  title={{The Persian lexicon project: minimized orthographic neighbourhood effects in a dense language}},
  author={Nemati, Fatemeh and Westbury, Chris and Hollis, Geoff and Haghbin, Hossein},
  journal={Journal of Psycholinguistic Research},
  volume={51},
  number={5},
  pages={957--979},
  year={2022},
  publisher={Springer}
}

@article{barrere2024training,
  title={Training transformer architectures on few annotated data: an application to historical handwritten text recognition},
  author={Barrere, Killian and Soullard, Yann and Lemaitre, Aur{\'e}lie and Co{\"u}asnon, Bertrand},
  journal={International Journal on Document Analysis and Recognition},
  volume={27},
  number={4},
  pages={553--566},
  year={2024},
  publisher={Springer Nature BV}
}
\bibliographystyle{elsarticle-num}

%
%
%
%




\end{document}